\documentclass[twoside,leqno,twocolumn]{article}

\usepackage[letterpaper]{geometry}
\usepackage{cite}
\usepackage{subfigure}
\usepackage{graphicx}
\usepackage{amsmath}
\usepackage{amssymb}
\usepackage{bm}
\usepackage{ltexpprt}
\usepackage{hyperref}
\usepackage{setspace}
\usepackage{multirow}
\usepackage{booktabs} 
\usepackage{tabularx}
\usepackage{longtable}
\usepackage{supertabular}
\usepackage{arydshln}
\usepackage[T1]{fontenc}
\usepackage{float}
\usepackage{stfloats}
\usepackage[algo2e, ruled,linesnumbered]{algorithm2e}
\usepackage{tabularray}
\usepackage{bbding}
\usepackage{caption}
\begin{document}
\begin{sloppypar}

% 1Code is available at https://github.com/mufusu21/MiceRec

%\setcounter{chapter}{2} % If you are doing your chapter as chapter one,
%\setcounter{section}{3} % comment these two lines out.

\title{\Large FedDCSR: Federated Cross-domain Sequential Recommendation via Disentangled Representation Learning}
% \author{Hongyu Zhang\thanks{Harbin Institute of Technology (Shenzhen)dd}
\author{
% Hongyu Zhang\thanks{Harbin Institute of Technology (Shenzhen)
Hongyu Zhang\thanks{Harbin Institute of Technology (Shenzhen), Shenzhen, China. orion-orion@stu.hit.edu.cn, xuyang97@stu.hit.edu.cn, fengjy@stu.hit.edu.cn, liaoqing@hit.edu.cn.}
\and Dongyi Zheng\thanks{Sun Yat-sen University, Guangzhou, China. zhengdy23@mail2.sysu.edu.cn.}
\and Xu Yang\footnotemark[1] 
\and Jiyuan Feng\footnotemark[1] 
% \and Xiangke Liao\footnotemark[2]
\and Qing Liao\footnotemark[1] \addtocounter{footnote}{2}\thanks{Corresponding author.}}

\date{}

\maketitle
% \fancyfoot[R]{\scriptsize{Copyright \textcopyright\ 2024 by SIAM\\
% Unauthorized reproduction of this article is prohibited}}

% Depending on which copyright you agree to when you sign the copyright form, the copyright
% can be changed to one of the following after commenting out the default copyright statement
% above.
%\pagenumbering{arabic}
%\setcounter{page}{1}%Leave this line commented out.

\begin{abstract} \small\baselineskip=9pt

% Cross-domain user sequential interactions can provide more powerful user representations than single-domain. 
Cross-domain Sequential Recommendation (CSR) which leverages user sequence data from multiple domains has received extensive attention in recent years. However, the existing CSR methods require sharing origin user data across domains, which violates the General Data Protection Regulation (GDPR). Thus, it is necessary to combine federated learning (FL) and CSR to fully utilize knowledge from different domains while preserving data privacy. Nonetheless, the sequence feature heterogeneity across different domains significantly impacts the overall performance of FL. In this paper, we propose \textbf{FedDCSR}, a novel federated cross-domain sequential recommendation framework via disentangled representation learning. Specifically, to address the sequence feature heterogeneity across domains, we introduce an approach called inter-intra domain sequence representation disentanglement (SRD) to disentangle the user sequence features into domain-shared and domain-exclusive features. In addition, we design an intra domain contrastive infomax (CIM) strategy to learn richer domain-exclusive features of users by performing data augmentation on user sequences. Extensive experiments on three real-world scenarios demonstrate that FedDCSR achieves significant improvements over existing baselines\footnote{Code available at \href{https://github.com/orion-orion/FedDCSR}{https://github.com/orion-orion/FedDCSR}}.
\end{abstract}

\section{Introduction}
In recent years, cross-domain sequential recommendation (CSR) has found wide applications in various scenarios such as e-commerce, social media, and video-sharing platforms. Based on the assumption that users have similar preferences across domains, CSR can significantly promote the next-item recommendation performance for users by leveraging user sequence data from multiple domains. However, existing works around CSR\cite{PINet, MIFN, C2DSR} require sharing origin data between domains, which violates the General Data Protection Regulation (GPDR). How to provide high-quality cross-domain sequential recommendations while preserving data privacy has become an urgent issue.

In this paper, we focus on a new paradigm called federated cross-domain sequential recommendation (FedCSR). In this context, user sequences are considered private information, which cannot be directly shared between domains. Although FedCSR can effectively solve the privacy issue in CSR, it also faces the challenge of sequence feature heterogeneity across domains, that is, user sequences in different domains contain domain-exclusive interaction information. 

\begin{figure}
\centering
\includegraphics[width=\linewidth, scale=1.00]{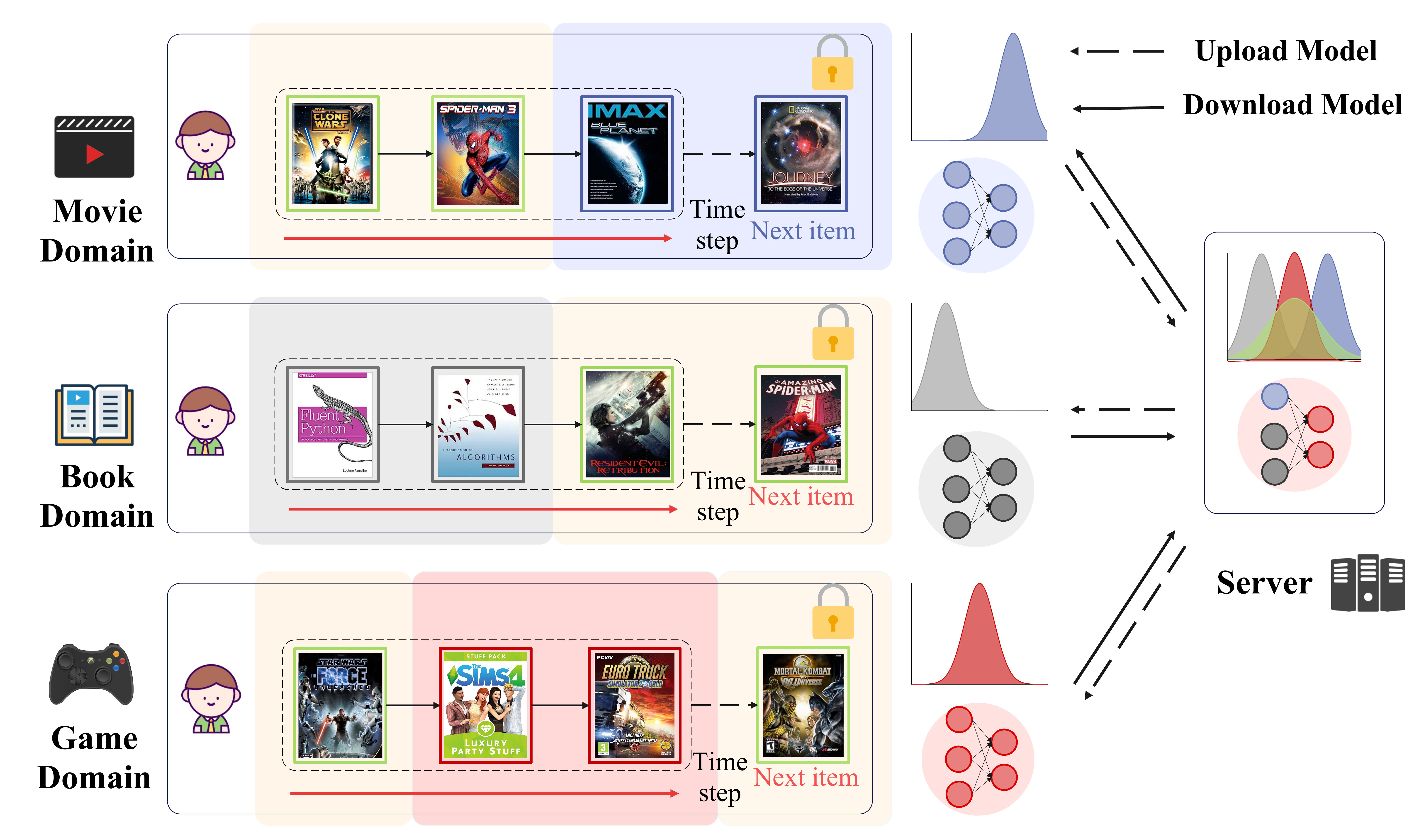}
\caption{Sequence feature heterogeneity across domains in the FedCSR scenario.}
% \end{center}
\label{fig1}
\end{figure}

It is impractical to directly apply classical federated learning methods such as FedAvg\cite{FedAvg} in the FedCSR scenario due to the sequence feature heterogeneity across domains. Specifically, Figure \ref{fig1} gives a toy example of the sequence feature heterogeneity across different domains. As shown in Figure\ref{fig1}, a user watches action movies and documentary movies in the Movie domain, reads professional books and action books in the Book domain, and plays action games and simulation games in the Game domain. As mentioned above, documentary movies in the Movie domain, professional books in the Book domain, and simulation games in the Game domain can all be regarded as domain-exclusive interaction information. Therefore, due to the presence of sequence feature heterogeneity, we can only obtain suboptimal results if all domains share the same model. 

To address the challenge above, we propose a novel federated cross-domain sequential recommendation framework via disentangled representation learning (FedDCSR), which allows different domains to train better-performing CSR models collaboratively without sharing origin user data. Specifically, inspired by the disentangled representation learning\cite{DSN, NIPS}, we introduce an inter-intra domain sequence representation disentanglement method called SRD to address the sequence feature heterogeneity across domains.  Following this method, the model of each domain is split into a local branch and a global branch, which are responsible for extracting domain-shared and domain-exclusive sequence representations separately. In addition, we design a contrastive infomax strategy CIM to learn richer domain-exclusive features of users by performing data augmentation on user sequences. We conduct the evaluation on Amazon datasets under the FL setting. The experimental results show that our FedDCSR can improve the performance of recommendations in three different CSR scenarios.

To summarize, our contributions are as follows:

\begin{itemize}
\item We propose a novel federated cross-domain sequential framework FedDCSR, which allows different domains to collaboratively train better performing CSR models, while effectively preserving data privacy.
\item We introduce SRD, an inter-intra domain sequence representation disentanglement method, which disentangles the user sequence features into domain-shared and domain-exclusive features to solve the sequence feature heterogeneity problem across domains.
\item We design an intra domain contrastive infomax strategy CIM, to learn richer domain-exclusive sequence features of users by performing data augmentation on user sequences.
\end{itemize}

% Our contributions are summarized as follows:

% \begin{itemize}
% \item We propose a novel federated learning framework FedDCSR, which enables various domains to train a more accurate cross-domain sequential recommendation model collaboratively without sharing origin data and effectively solves the problem of domain feature heterogeneity;
% \item We propose a feature modeling method based on disentangled learning, which splits the local model into a local branch and a global branch and disentangles the two sequence features of users to solve the feature heterogeneity problem;
% \item We design a contrastive infomax module to learn more effective user features by performing data augmentation and feature comparison on user sequences.
% \end{itemize}

\section{Related Work.}

% We consider three types of related work: sequential, cross-domain sequential, and federated cross-domain recommendation.  

\textbf{Sequential Recommendation} SASRec\cite{SASRec} first uses the self-attention network to model user sequences. VSAN\cite{VSAN} introduces variational inference into self-attention networks for sequential recommendation. ContrastVAE\cite{ContrastVAE} proposes to use contrastive learning to solve the posterior collapse problem in VAE. CL4SRec\cite{CL4SRec} and  DuoRec\cite{DuoRec} introduce the contrastive learning framework to extract self-supervised signals from user sequences. However, the above methods only focus on a single domain and cannot fully utilize user data from multiple domains.

\textbf{Cross-domain Recommendation} DDTCDR\cite{DDTCDR} proposes a dual transfer learning based model that improves recommendation performance across domains. DisAlign\cite{DisAlign} proposes Stein path alignment to align the distributions of embeddings across domains. CDRIB\cite{CDRIB} proposes information bottleneck regularizers to build user-item correlations across domains. However, the above methods should access the whole user-item interactions across domains and are not feasible in the FL setting. 

\textbf{Cross-domain Sequential Recommendation} As a pioneer work, $\pi$-net \cite{PINet} utilizes a gating mechanism to enhance the domain knowledge transfer. Based on this, MIFN\cite{MIFN} further introduces knowledge graphs to build bridges between different domains. C$^2$DSR \cite{C2DSR} proposes a contrastive infomax objective for modeling the relationship between domains. While these methods still face the same problems as above.

% However, the above methods need to access the whole user sequences across domains and are not feasible in the FL setting. 

\textbf{Federated Cross-domain Recommendation} FedCT\cite{FedCT} employs variational autoencoders to generate shared encoding latent vectors across domains. FedCDR\cite{FedCDR} sets up the user personalization model on the client side and uploads other models to the server during aggregation. FedCTR\cite{FedCTR} designs a framework to train a CTR prediction model across multiple platforms under privacy protection. However, none of them have considered the sequential data and addressed the sequence feature heterogeneity problem across domains.

\begin{figure*}
% \vspace{-3em}
\centering
\includegraphics[height=0.55\linewidth, scale=0.7]{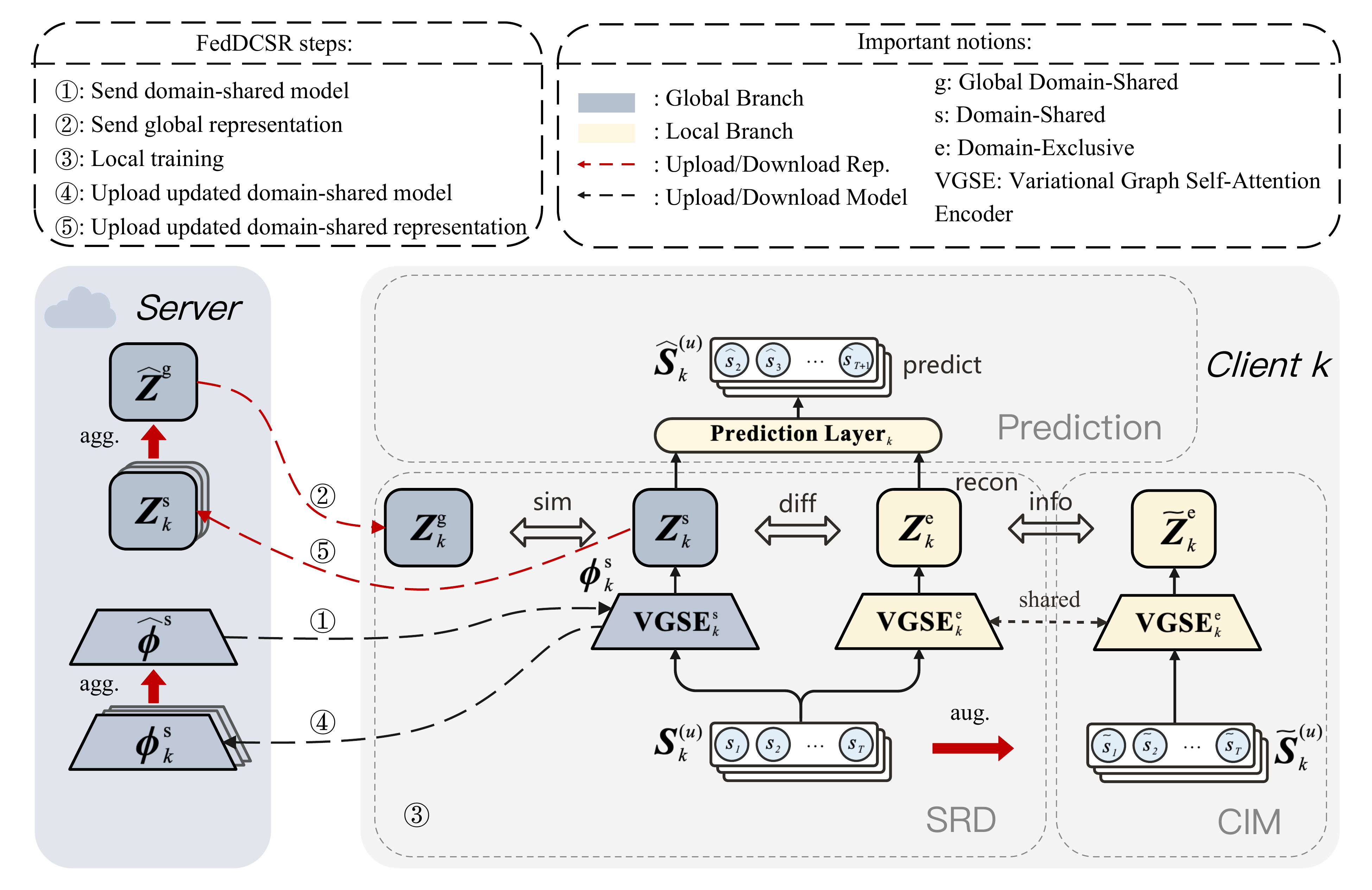}
\caption{An overview of FedDCSR.}
% \end{center}
\label{FedDCSR-framework}
\end{figure*}

\section{Methodology.}

% In this section, we describe our proposed federated cross-domain sequential recommendation framework via disentangled learning(FedDCSR). In section 3.1, we first describe our problem definition. In section 3.2, we introduce the overall idea and the framework of FedDCSR. In section 3.3, we introduce the proposed variational graph self-attention encoder(VGSE). In section 3.4, we introduce the proposed intra-inter domain representation disentanglement objective, which includes three parts: difference loss, similarity loss, and intra domain reconstruction loss. In section 3.5, we introduce the proposed intra domain contrastive infomax loss, which uses contrastive information maximization to enhance the user preference information contained in the domain exclusive representation. In section 3.6, we introduce the federated training and evaluation procedure. 
% In section 3.6, we present an algorithm and its technical components in detail

\subsection{Problem Definition.}

Assume there are $K$ local clients and a central server. The $k$-th client maintains its own user sequence dataset denoted as $\mathcal{D}_k=\{\boldsymbol{S}^{(u)}_k\}_{u\in \mathcal{U}_k}$, which forms a distinct domain, where $\mathcal{U}_k$ is the user set in client $k$ and $\boldsymbol{S}^{(u)}_k$ is the user sequence for a certain user $u$. The sequence $\boldsymbol{S}_k^{(u)}=(s_1, s_2, \cdots, s_T)$ contains the interactions between user $u$ and items within the past $T$ time steps. 

For client $k$, we disentangle the user sequence feature into domain-shared representations $\boldsymbol{Z}^{\mathrm{s}}_k$ and domain-exclusive representations $\boldsymbol{Z}^{\mathrm{e}}_k$. After the local model update is completed, the central server aggregates ${\{\boldsymbol{Z}^{\mathrm{s}}_k\}}^K_{k=1}$ to obtain the global representations $\boldsymbol{Z}^{\mathrm{g}}$ used in the next training round. The local augmented user representations is denoted as $\widetilde{\boldsymbol{Z}_k}^{\mathrm{e}}$. Each client's local model is split into a global branch including a domain-shared encoder (parameterized by $\boldsymbol{\phi}^{\mathrm{s}}_k$), and a local branch including a domain-exclusive encoder (parameterized by $\boldsymbol{\phi}^{\mathrm{e}}_k$) and a prediction layer (parameterized by $\boldsymbol{\theta}_k$). After each round of training is completed, the server aggregates $\{\boldsymbol{\phi}_k^{\mathrm{s}}\}^{K}_{k=1}$ to get $\boldsymbol{\phi}^{\mathrm{g}}$ which will be shared among clients in the next training round.

\subsection{Overview of FedDCSR.}

As shown in Figure\ref{FedDCSR-framework}, our proposed FedDCSR adopts client-server federated learning architecture. The model of each client is split into a local branch (in yellow) and a global branch (in purple). Only domain-shared representations and model parameters are aggregated in each training round. During the test stage, domain-shared and domain-exclusive representations are jointly employed for local prediction.

% First, the server sends the latest domain-shared model to different participating clients. Then each client $k$ uses its local dataset $\mathcal{D}_k$ to compute difference loss, reconstruction loss, and contrastive infomax loss(note that the global representations $\boldsymbol{Z}_g$ has not been obtained in the first round of iterations, so the similarity loss is not computed), and perform local model and representation updates. After the local update is complete, clients send the parameters of the updated domain-shared model and domain-shared representations to the server. Then the server aggregates domain-shared models to generate the latest domain-shared model, and aggregates domain-shared representations to generate the latest global representations. In the next iteration, the server sends the domain-shared model and domain-shared representations to the clients. This process is repeated until the local models achieve convergence.

\subsection{Variational Graph Self-Attention Encoder (VGSE).}

In this section, we introduce our proposed VGSE, which is used to model the representation distribution of local user sequences, taking into account the relationships between items.

Assume that user sequences in the local dataset $\mathcal{D}_k={\{\boldsymbol{S}_k^{(u)}\}}_{u\in \mathcal{U}_k}$ are drawn from the following distribution:
\begin{equation}
\begin{aligned}
p_{\boldsymbol{\theta}_k}\left(\boldsymbol{S}_k\right)=\int p_{\boldsymbol{\theta_k}}\left(\boldsymbol{S}_k \mid \boldsymbol{Z}_k^{\mathrm{s}}, \boldsymbol{Z}_k^{\mathrm{e}}\right)p(\boldsymbol{Z}_k^{\mathrm{s}})p( \boldsymbol{Z}_k^{\mathrm{e}}) d\boldsymbol{Z}_k^{\mathrm{s}}d\boldsymbol{Z}_k^{\mathrm{e}},
\end{aligned}
\label{joint}
\end{equation}
where we assume the prior distributions $p(\boldsymbol{Z}_k^{\mathrm{s}})$, $p(\boldsymbol{Z}_k^{\mathrm{e}})$ following normal Gaussian  distribution $\mathcal{N}(\boldsymbol{0}, \boldsymbol{I})$, and the conditional distribution of $\boldsymbol{S}_k$ is defined as follows:
\begin{equation}
\vspace{-0.15cm}
\resizebox{.99\hsize}{!}{
\begin{math}
\begin{aligned}
p_{\boldsymbol{\theta}_{k}}\left(\boldsymbol{S}_k \mid \boldsymbol{Z}_k^{\mathrm{s}}, \boldsymbol{Z}_k^{\mathrm{e}}\right) & =\prod_{t=2}^T p_{\boldsymbol{\theta}_{k}}\left(s_t \mid s_1, \cdots, s_{t-1}, \boldsymbol{Z}_k^{\mathrm{s}}, \boldsymbol{Z}_k^{\mathrm{e}} \right) \\
& =\prod_{t=2}^T p_{\boldsymbol{\theta}_{k}}\left(s_t \mid \boldsymbol{z}_k^{\mathrm{s}, 1}, \cdots \boldsymbol{z}_k^{\mathrm{s}, t-1}, \boldsymbol{z}_k^{\mathrm{e}, 1}, \cdots \boldsymbol{z}_k^{\mathrm{e}, t-1}\right),
\end{aligned}
\end{math}
}
\end{equation}
where $\boldsymbol{Z}^{\mathrm{s}}_k, \boldsymbol{Z}^{\mathrm{e}}_k \in \mathbb{R}^{T\times d}$, and $\boldsymbol{z}_{k}^{\mathrm{s}, t}, \boldsymbol{z}_{k}^{\mathrm{e}, t} \in \mathbb{R}^{d}, t=[1,2, \ldots, T]$ denote the representation of the user sequence at the $t$-th time step.

Because the integral of the marginal likelihood in (\ref{joint}) is intractable and the true posterior $p_{\boldsymbol{\theta}_k}(\boldsymbol{Z}^{\mathrm{s}}_k, \boldsymbol{Z}^{\mathrm{e}}_k\mid \boldsymbol{S}_k)$ is intractable as well, we use the approximated posterior $q_{\boldsymbol{\phi}_k}(\boldsymbol{Z}^{\mathrm{s}}_k,\boldsymbol{Z}^{\mathrm{e}}_k\mid \boldsymbol{S}_k)$, which is coefficientized according to the probabilistic graphical model in Figure\ref{PGM} as follows:  
\begin{equation}
q_{\boldsymbol{\phi}_k}(\boldsymbol{Z}^{\mathrm{s}}_k,\boldsymbol{Z}^{\mathrm{e}}_k\mid \boldsymbol{S}_k) = q_{\boldsymbol{\phi}_k^{\mathrm{s}}}(\boldsymbol{Z}^{\mathrm{s}}_k\mid \boldsymbol{S}_k)q_{\boldsymbol{\phi}_k^{\mathrm{e}}}(\boldsymbol{Z}^{\mathrm{e}}_k\mid \boldsymbol{S}_k),
\end{equation}
where $q_{\boldsymbol{\phi}_k^{\mathrm{s}}}(\boldsymbol{Z}^{\mathrm{s}}_k\mid \boldsymbol{S}_k)$ and $q_{\boldsymbol{\phi}_k^{\mathrm{e}}}(\boldsymbol{Z}^{\mathrm{e}}_k\mid \boldsymbol{S}_k)$ are domain-shared and domain-exclusive VGSE respectively. 

\begin{figure}
% \vspace{-3em}
\centering
\includegraphics[scale=0.15]{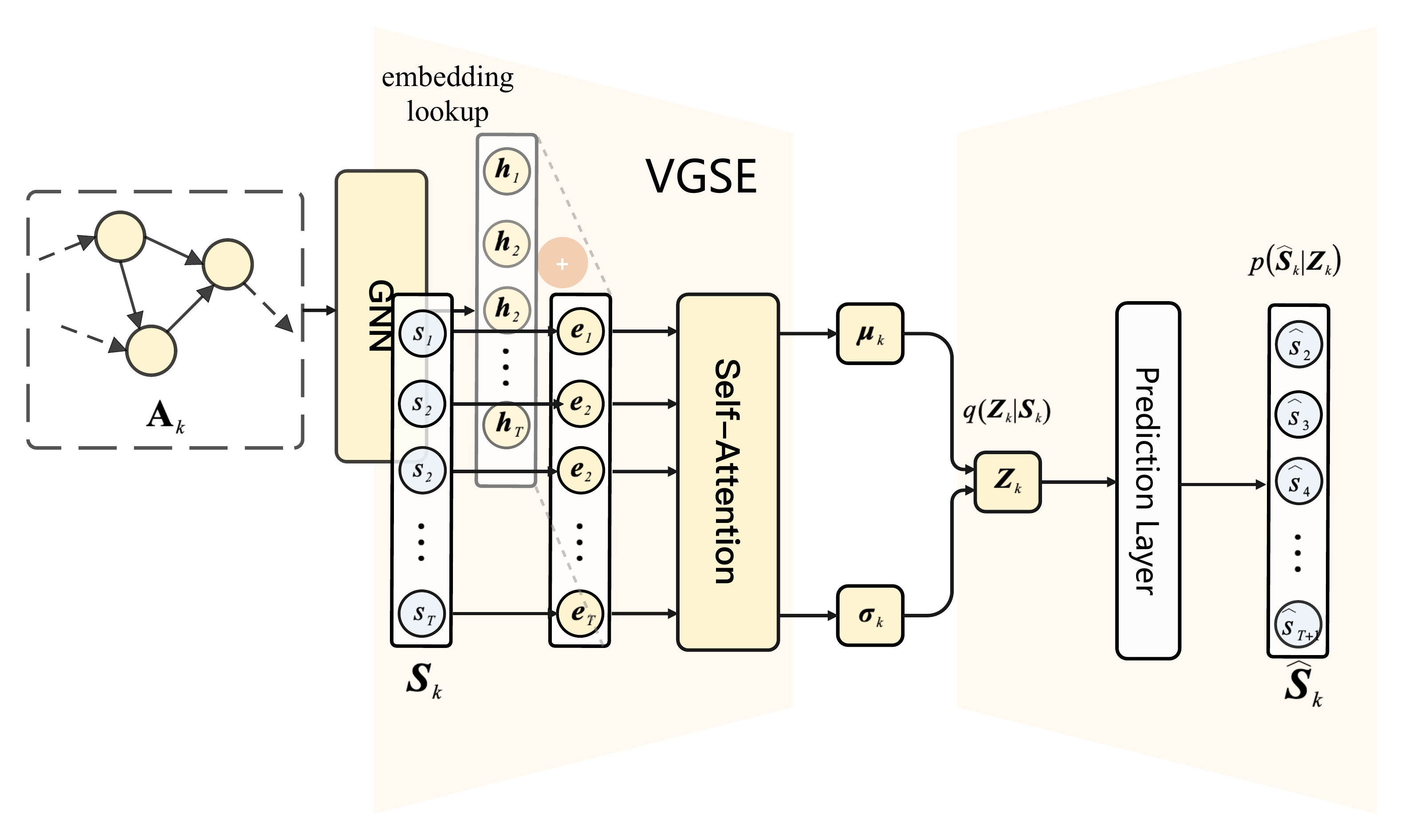}
\caption{The architecture of the variational graph self-attention encoder.}
% \end{center}
\label{VGSE}
\end{figure}

\begin{figure}
\centering
% \vspace{-3em}
\includegraphics[scale=0.35]{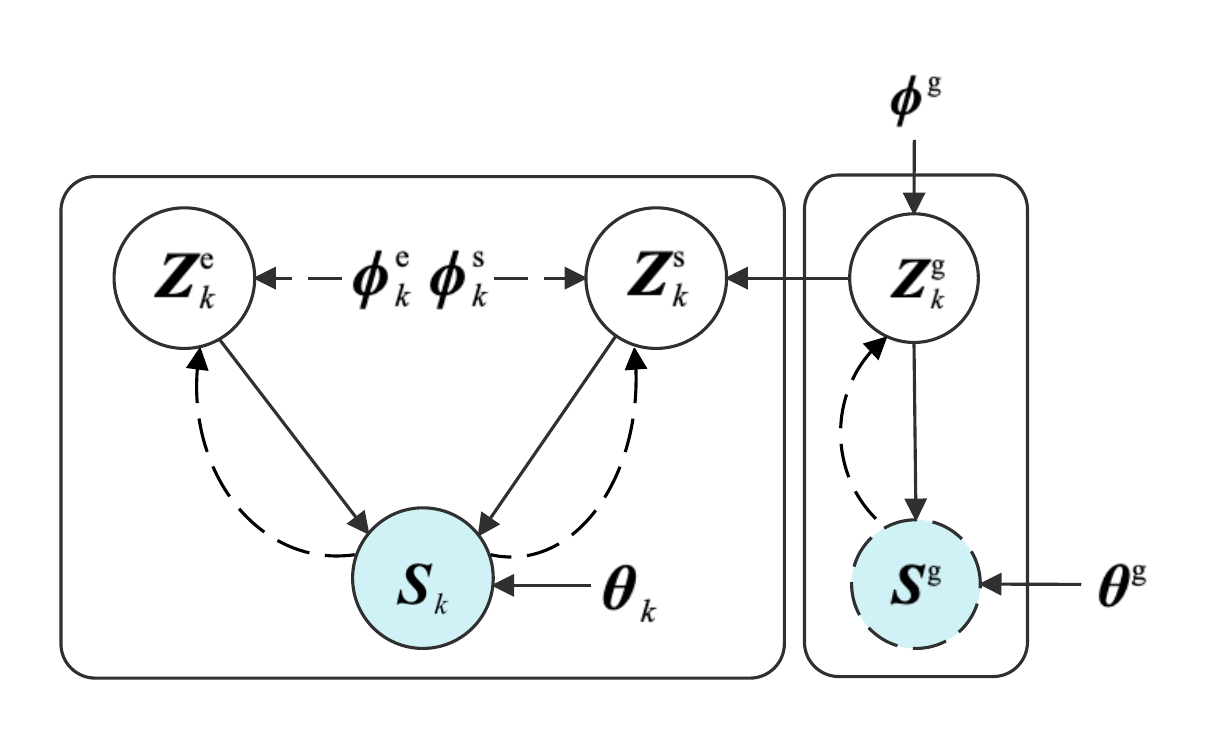}
\caption{The graphical model illustrating the relationship between  $\boldsymbol{Z}^{\mathrm{s}}_k$, $\boldsymbol{Z}^{\mathrm{e}}_k$, $\boldsymbol{Z}^{\text{g}}_k$, $\boldsymbol{S}_k$ and the dummy variable $\boldsymbol{S}^{\mathrm{g}}$, which follows the global squence distribution.}
% \end{center}
\label{PGM}
\end{figure}

Next, we will illustrate how VGSE is designed. As shown in Figure \ref{VGSE}, we first process the item relationship matrix $\boldsymbol{A}_k$ through two GNNs to obtain the domain-shared and domain-exclusive embeddings of items. The update formula of the  $l$-th GNN layer (totally $L$ layer) is defined as follows:
\begin{equation}
\begin{aligned}
\boldsymbol{H}_k^{\mathrm{s}, l}=\mathrm{Norm}\left(\boldsymbol{A}_k\right) \boldsymbol{H}_k^{\mathrm{s}, l-1}, 
\quad \boldsymbol{H}_k^{\mathrm{e}, l}=\mathrm{Norm}\left(\boldsymbol{A}_k\right) \boldsymbol{H}_k^{\mathrm{e}, l-1},
\end{aligned}
\end{equation}
where $l=1,2, \cdots, L$. Let $\boldsymbol{H}_k^{\mathrm{s}}, \boldsymbol{H}_k^{\mathrm{e}}$ denote the average embeddings of the items obtained in all layers of GNN. Then for input sequence $\boldsymbol{S}_k$, we obtain its corresponding domain-shared and domain-exclusive relationship embeddings $\boldsymbol{H}_{k, s_k}^{\mathrm{s}}, \boldsymbol{H}_{k, s_k}^{\mathrm{e}}$ through the $\mathrm{EmbeddingLookup}$ operation. Let $\boldsymbol{E}_k^{\mathrm{s}}, \boldsymbol{E}_k^{\mathrm{e}}$ denote the sequential embeddings of the sequence, and $\boldsymbol{P}_k^{\mathrm{s}}, \boldsymbol{P}_k^{\mathrm{e}}$ denote the positional embeddings of the sequence. Then we add up $\boldsymbol{H}_{k, s_k}^{\text{s}}$, $\boldsymbol{P}_k^{\text{s}}$, $\boldsymbol{E}_k^{\text{s}}$ and $\boldsymbol{H}_{k, s_k}^{\text{e}}$, $\boldsymbol{P}_k^{\text{e}}$, $\boldsymbol{E}_k^{\text{e}}$ separately, then feed these sums into self-attention networks to get attention aggregation results $\boldsymbol{\mu}_k^{\mathrm{s}}$, $\boldsymbol{\sigma}_k^{\mathrm{s}}$ and $\boldsymbol{\mu}_k^{\mathrm{e}}$, $\boldsymbol{\sigma}_k^{\mathrm{e}}$:
\begin{equation}
\begin{aligned}
& \boldsymbol{\mu}_k^{\mathrm{s}},  \boldsymbol{\sigma}_k^{\mathrm{s}}=\text { Self-Attention}^{\mathrm{s}}_k\left(\boldsymbol{E}_k^{\mathrm{s}}+\boldsymbol{P}_k^{\mathrm{s}} +\boldsymbol{H}_{k, s_k}^{\mathrm{s}}\right), \\
& \boldsymbol{\mu}_k^{\mathrm{e}},  \boldsymbol{\sigma}_k^{\mathrm{e}}=\text { Self-Attention}^{\mathrm{e}}_k \left(\boldsymbol{E}_k^{\mathrm{e}}+\boldsymbol{P}_k^{\mathrm{e}}+\boldsymbol{H}_{k, s_k}^{\mathrm{e}}\right). \\
\end{aligned}
\end{equation}
Next, we use the reparameterization trick to sample the domain-shared and the domain-exclusive representations $\boldsymbol{Z}_k^{\mathrm{s}}, \boldsymbol{Z}_k^{\mathrm{e}}$ from $q_{\boldsymbol{\phi}_k^{\mathrm{s}}}(\boldsymbol{Z}^{\mathrm{s}}_k\mid \boldsymbol{S}_k)$ and $q_{\boldsymbol{\phi}_k^{\mathrm{e}}}(\boldsymbol{Z}^{\mathrm{e}}_k\mid \boldsymbol{S}_k)$:
\begin{equation}
\boldsymbol{Z}_k^{\mathrm{s}}=\boldsymbol{\mu}_k^{\mathrm{s}}+\boldsymbol{\sigma}_k^{\mathrm{s}} \odot \boldsymbol{\epsilon}, \quad\boldsymbol{Z}_k^{\mathrm{e}}=\boldsymbol{\mu}_k^{\mathrm{e}} +\boldsymbol{\sigma}_k^{\mathrm{e}} \odot \boldsymbol{\epsilon},
\end{equation}
where $\epsilon \sim \mathcal{N} (\mathbf{0}, \boldsymbol{I}) $ is Gaussian noise.

\subsection{Intra-Inter Domain Sequence Representation Disentanglement (SRD).}

In this section, we introduce SRD, which disentangles user sequence representations into domain-shared and domain-exclusive representations. Specifically, SRD includes three components: difference loss, similarity loss, and reconstruction loss.

\subsubsection{Difference Loss} The difference loss is applied to each domain and encourages the domain-shared and domain-exclusive encoders to encode distinct information of the inputs. The structured probabilistic relationship between $\boldsymbol{Z}^{\mathrm{s}}_k$, $\boldsymbol{Z}^{\mathrm{e}}_k$ and $\boldsymbol{S}_k$ is shown in Figure \ref{VGSE}. Then the difference loss $\mathcal{L}_k^{\text {diff }}$ during local training can be formulated as:
\begin{equation}
\begin{aligned}
\mathcal{L}_k^{\text {diff }}=I\left(\boldsymbol{Z}_k^{\mathrm{s}} ; \boldsymbol{Z}_k^{\mathrm{e}}\right),
\end{aligned}
\end{equation}
Here $I(\cdot, \cdot)$ denotes the mutual information of two random variables. Intuitively, The above formula aims to amplify the difference between domain-shared representations and domain-exclusive representations.

To minimize $I\left(\boldsymbol{Z}_k^{\mathrm{s}} ; \boldsymbol{Z}_k^{\mathrm{e}}\right)$, we need to make a further equivalent transformation. Following the definition of interaction information\cite{MMI}, we have:
\begin{equation}
\begin{aligned}
I\left(\boldsymbol{Z}_k^{\mathrm{s}} ; \boldsymbol{Z}_k^{\mathrm{e}}\right) & =I\left(\boldsymbol{Z}_k^{\mathrm{s}} ; \boldsymbol{Z}_k^{\mathrm{e}}; \boldsymbol{S}_k\right) + I(\boldsymbol{Z}^{\mathrm{s}}_k; \boldsymbol{Z}^{\mathrm{e}}_k \mid \boldsymbol{S}_k).\\
\end{aligned}
\label{interaction information}
\end{equation}
According to the structural assumption in Figure \ref{PGM}, we have $q(\boldsymbol{Z}^{\mathrm{s}}_k \mid \boldsymbol{S}_k) = q(\boldsymbol{Z}^{\mathrm{s}}_k \mid \boldsymbol{Z}^{\mathrm{e}}_k,  \boldsymbol{S}_k)$, thus the last term $
I(\boldsymbol{Z}^{\mathrm{s}}_k; \boldsymbol{Z}^{\mathrm{e}}_k \mid \boldsymbol{S}_k) = H(\boldsymbol{Z}^{\mathrm{s}}_k \mid \boldsymbol{S}_k) - H(\boldsymbol{Z}^{\mathrm{s}}_k \mid \boldsymbol{Z}^{\mathrm{e}}_k,  \boldsymbol{S}_k) = 0$, where $H(\cdot)$ denotes the information entropy. Then we have
\begin{equation}
\begin{aligned}
I\left(\boldsymbol{Z}_k^{\mathrm{s}} ; \boldsymbol{Z}_k^{\mathrm{e}}\right)
& =I\left(\boldsymbol{Z}_k^{\mathrm{s}} ; \boldsymbol{Z}_k^{\mathrm{e}} ; \boldsymbol{S}_k\right) \\
& =I\left(\boldsymbol{S}_k ; \boldsymbol{Z}_k^{\mathrm{s}}\right)-I\left(\boldsymbol{S}_k ; \boldsymbol{Z}_k^{\mathrm{s}} \mid \boldsymbol{Z}_k^{\mathrm{e}}\right) \\
& =I\left(\boldsymbol{S}_k ; \boldsymbol{Z}_k^{\mathrm{s}}\right)+I\left(\boldsymbol{S}_k ; \boldsymbol{Z}_k^{\mathrm{e}}\right)-I\left(\boldsymbol{S}_k ; \boldsymbol{Z}_k^{\mathrm{s}}, \boldsymbol{Z}_k^{\mathrm{e}}\right).
\end{aligned}
\end{equation}
However, directly minimizing the above formula is still intractable, so we need to minimize its variational upper bound \cite{NIPS} instead:
\begin{equation}
\begin{aligned}
I\left(\boldsymbol{Z}_k^{\mathrm{s}} ; \boldsymbol{Z}_k^{\mathrm{e}}\right) \leq & D_{\mathrm{KL}}\left(q\left(\boldsymbol{Z}_k^{\mathrm{s}} \mid \boldsymbol{S}_k\right) \| p\left(\boldsymbol{Z}_k^{\mathrm{s}}\right)\right)\\
&+D_{\mathrm{KL}}\left(q\left(\boldsymbol{Z}_k^{\mathrm{e}} \mid \boldsymbol{S}_k\right) \| p\left(\boldsymbol{Z}_k^{\mathrm{e}}\right)\right) \\
& -\mathbb{E}_{q\left(\boldsymbol{Z}_k^{\mathrm{s}} \mid \boldsymbol{S}_k\right)q\left(\boldsymbol{Z}_k^{\mathrm{e}} \mid \boldsymbol{S}_k\right)}\left[\log p\left(\boldsymbol{S}_k \mid \boldsymbol{Z}_k^{\mathrm{s}}, \boldsymbol{Z}_k^{\mathrm{e}}\right)\right].
\end{aligned}
\end{equation}
Intuitively, the above formula is to align the posterior distribution $q\left(\boldsymbol{Z}_k^{\mathrm{s}} \mid \boldsymbol{S}_k\right)$ and $q\left(\boldsymbol{Z}_k^{\mathrm{e}} \mid \boldsymbol{S}_k\right)$ with the Gaussian distribution, and enable $\boldsymbol{Z}_k^{\mathrm{s}}$ and $\boldsymbol{Z}_k^{\mathrm{e}}$ can jointly reconstruct $\boldsymbol{S}_k$.

\subsubsection{Similarity Loss} 
The similarity loss of each domain encourages the domain-shared encoders to encode the information shared between domains. For client $k$, The structured probabilistic relationship between $\boldsymbol{Z}^{\mathrm{s}}_k$ and $\boldsymbol{Z}^{\mathrm{g}}_k$ is shown in Figure \ref{PGM}. Then the similarity loss $\mathcal{L}_k^{\operatorname{sim}}$ during local training can be defined as follows:
\begin{equation}
\begin{aligned}
\mathcal{L}_k^{\operatorname{sim}}=-I\left(\boldsymbol{Z}_k^{\mathrm{s}} ; \boldsymbol{Z}^{\mathrm{g}}_k\right).
\end{aligned}
\end{equation}
Intuitively, the purpose of the above formula is to make the domain-shared representations and global representations similar to facilitate knowledge transfer.

As mentioned above, directly maximizing $I\left(\boldsymbol{Z}_k^{\mathrm{s}} ; \boldsymbol{Z}^{\mathrm{g}}\right)$ is intractable. We adopt the idea of Deep Infomax\cite{MINE, DIM} to use Jensen-Shannon Divergence (JSD) as its variational lower bound:
\begin{equation}
\begin{aligned}
I\left(\boldsymbol{Z}_k^{\mathrm{s}} ; \boldsymbol{Z}^{\text{g}}_k\right) &\geq \hat{I}^{\mathrm{JSD}}\left(\boldsymbol{Z}_k^{\mathrm{s}} ; \boldsymbol{Z}^{\text{g}}_k\right) \\
 &= \mathbb{E}_{q\left(\boldsymbol{Z}^{\mathrm{s}}_k, \boldsymbol{Z}^{\mathrm{g}}_k \mid \boldsymbol{S}_k, \boldsymbol{S}^{\mathrm{g}}\right)}\left[-\mathrm{sp}\left(-T_{k}\left(\boldsymbol{Z}^{\mathrm{s}}_k, \boldsymbol{Z}^{\mathrm{g}}_k\right)\right)\right] \\
 &\quad- \mathbb{E}_{q\left(\boldsymbol{Z}_k^{\mathrm{s}} \mid \boldsymbol{S}_k\right)q\left(\boldsymbol{Z}^{\mathrm{g}} \mid \boldsymbol{S}^{\mathrm{g}}\right)} \left[\mathrm{sp}(T_{k}(\boldsymbol{Z}^{\mathrm{s},\prime}_k, \boldsymbol{Z}^{\mathrm{g}}_k))\right]. \\
% & =\sum_{u_i \in \mathcal{U}_k, u_j \in \mathcal{U}^{\text{g}}}\left[\log \left(D_k\left(\boldsymbol{Z}_{k, u_i}^{\mathrm{s}}, \boldsymbol{Z}_{u_j}^{\text{g}}\right)\right)\right.\\
% &\quad\quad\quad\quad\quad\text{   } + \log \left(1-D_k\left(\boldsymbol{Z}_{k, u_i}^{\mathrm{s}\text{ }\prime}, \boldsymbol{Z}_{u_j}^{\text{g}}\right)\right)]
\end{aligned}
% \geq \mathbb{E}_{q_\phi\left(\boldsymbol{Z}_k^{\mathrm{s}} \mid \boldsymbol{S}_k\right), q_\phi\left(\boldsymbol{Z}^{\text{g}} \mid \boldsymbol{S}^{\text{g}}\right)}\left[\log D_k\left(\boldsymbol{Z}_k^{\mathrm{s}}, \boldsymbol{Z}^{\text{g}}\right)\right] \\
% & =\sum_{u_i \in \mathcal{U}_k, u_j \in \mathcal{U}^{\text{g}}}\left[\log \left(D_k\left(\boldsymbol{Z}_{k, u_i}^{\mathrm{s}}, \boldsymbol{Z}_{u_j}^{\text{g}}\right)\right)\right.\\
% &\quad\quad\quad\quad\quad\text{   } + \log \left(1-D_k\left(\boldsymbol{Z}_{k, u_i}^{\mathrm{s}\text{ }\prime}, \boldsymbol{Z}_{u_j}^{\text{g}}\right)\right)]
\end{equation}
In the above formula, the discriminator $T_k$ differentiates between a paired sample from the joint (positive pair) and a paired sample from the product of marginals (negative pair), $\mathrm{sp}(z) = \log(1+e^z)$ is the softplus function, and $\boldsymbol{Z}_{k}^{\mathrm{s}\text{ }\prime}$ denotes negative sample latent vectors, here we set $\boldsymbol{Z}_{k}^{\mathrm{s}\text{ }\prime} = \boldsymbol{Z}_{k}^{\mathrm{e}}$ for each user. This bound enables the estimation of mutual information with only a single negative example for each positive example.

\subsubsection{Reconstruction Loss}
Furthermore, it is crucial to empower domain-exclusive representations with the capability to reconstruct sequential data in their respective domains in order to minimize the dilution of domain-exclusive information. The structural probabilistic relationship between $\boldsymbol{Z}_k^{\mathrm{e}}$ and $\boldsymbol{S}_k$ is shown in Figure \ref{VGSE}. Then the reconstruction loss $\mathcal{L}_k^{\text {recon }}$ during training can be formulated as:
\begin{equation}
\begin{aligned}
\mathcal{L}_k^{\text {recon }}=-I\left(\boldsymbol{Z}_k^{\mathrm{e}} ; \boldsymbol{S}_k\right).
\end{aligned}
\end{equation}
Intuitively, the above formula correlates the domain-exclusive representations $\boldsymbol{Z}^{\mathrm{e}}_k$ with the user sequence $\boldsymbol{S}_k$ to complete the reconstruction of intra domain information.

As before, to maximize $I\left(\boldsymbol{Z}_k^{\mathrm{e}} ; \boldsymbol{S}_k\right)$, we need to maximize its variational lower bound\cite{VMI}:
\begin{equation}
\begin{aligned}
I\left(\boldsymbol{Z}_k^{\mathrm{e}} ; \boldsymbol{S}_k\right) & =\mathbb{E}_{p\left(\boldsymbol{S}_k\right)q\left(\boldsymbol{Z}_k^{\mathrm{e}} \mid \boldsymbol{S}_k\right)}\left[\log \frac{p\left(\boldsymbol{S}_k \mid \boldsymbol{Z}_k^{\mathrm{e}}\right)}{p\left(\boldsymbol{S}_k\right)}\right] \\
& \geq \mathbb{E}_{p\left(\boldsymbol{S}_k\right)q\left(\boldsymbol{Z}_k^{\mathrm{e}} \mid \boldsymbol{S}_k\right)}\left[\log p\left(\boldsymbol{S}_k \mid \boldsymbol{Z}_k^{\mathrm{e}}\right)\right] + H(\boldsymbol{S}_k)\vspace{-3.5em}.
\end{aligned}
% \vspace{-1em}
\end{equation}

% The last term $\mathcal{H}(\boldsymbol{S}_k)$ only involves the generation process of sequence data, and has no connection with the model, 

% and the former term also fits $p(x|zx, zs)$ to  $q(x|zx, zs)$ so that we can utilize it as a decoder.

Afterward, adding up the difference loss, the similarity loss, and the reconstruction loss, the intra-inter domain representation disentanglement loss can be derived as:
\begin{equation}
\setlength{\belowdisplayskip}{8pt}
\begin{aligned}
\mathcal{L}_k^{\text {disen }} & =\alpha \mathcal{L}_k^{\text {diff }}+\beta \mathcal{L}_k^{\text {sim }}+\gamma \mathcal{L}_k^{\text {recon }} \\
& =\alpha I\left(\boldsymbol{Z}_k^{\mathrm{s}} ; \boldsymbol{Z}_k^{\mathrm{e}}\right)-\beta I\left(\boldsymbol{Z}_k^{\mathrm{s}} ; \boldsymbol{Z}^{\mathrm{g}}_k\right)-\gamma I\left(\boldsymbol{Z}_k^{\mathrm{e}} ; \boldsymbol{S}_k\right) \\
& \leq \alpha D_{\mathrm{KL}}\left(q\left(\boldsymbol{Z}_k^{\mathrm{s}} \mid \boldsymbol{S}_k\right) \| p\left(\boldsymbol{Z}_k^{\mathrm{s}}\right)\right) \\
& \quad +\alpha D_{\mathrm{KL}}\left(q\left(\boldsymbol{Z}_k^{\mathrm{e}} \mid \boldsymbol{S}_k\right) \| p\left(\boldsymbol{Z}_k^{\mathrm{e}}\right)\right) \\
& \quad -\alpha \mathbb{E}_{q\left(\boldsymbol{Z}_k^{\mathrm{s}} \mid \boldsymbol{S}_k\right)q\left(\boldsymbol{Z}_k^{\mathrm{e}} \mid \boldsymbol{S}_k\right)}\left[\log p\left(\boldsymbol{S}_k \mid \boldsymbol{Z}_k^{\mathrm{s}}, \boldsymbol{Z}_k^{\mathrm{e}}\right)\right] \\
& \quad - \beta \mathbb{E}_{q\left(\boldsymbol{Z}^{\mathrm{s}}_k, \boldsymbol{Z}^{\mathrm{g}}_k \mid \boldsymbol{S}_k, \boldsymbol{S}^{\mathrm{g}}\right)}\left[-\mathrm{sp}\left(-T_{k}\left(\boldsymbol{Z}^{\mathrm{s}}_k, \boldsymbol{Z}^{\mathrm{g}}_k\right)\right)\right] \\
& \quad + \beta \mathbb{E}_{q\left(\boldsymbol{Z}_k^{\mathrm{s}} \mid \boldsymbol{S}_k\right)q\left(\boldsymbol{Z}^{\mathrm{g}}_k \mid \boldsymbol{S}^{\mathrm{g}}\right)} \left[\mathrm{sp}(T_{k}(\boldsymbol{Z}^{\mathrm{s},\prime}_k, \boldsymbol{Z}^{\mathrm{g}}_k))\right] \\
& \quad -\gamma \mathbb{E}_{q\left(\boldsymbol{Z}_k^{\mathrm{s}} \mid \boldsymbol{S}_k\right)}\left[\log p\left(\boldsymbol{S}_k \mid \boldsymbol{Z}_k^{\mathrm{s}}\right)\right],
\end{aligned}
\end{equation}
where $\alpha$, $\beta$, and $\gamma$ are hyperparameters that control the degree of intra-inter domain representation disentanglement.

\subsection{Intra Domain Contrastive Infomax (CIM).}
In this section, we introduce CIM, which uses contrastive information maximization\cite{SimCLR, InfoNCE} to enhance the user preference information contained in the domain-exclusive representations $\boldsymbol{Z}_k^{\mathrm{e}}$.

In client $k$, let us denote the user sequence after data augmentation as $\widetilde{\boldsymbol{S}}_k$. The data augmentation method employed in this study is randomly shuffling the sequence of user sequences, ie., $\widetilde{\boldsymbol{S}}_k=\left(\widetilde{s}_1, \widetilde{s}_2, \cdots, \widetilde{s}_T\right)$. Then let $\widetilde{\boldsymbol{Z}}^{\mathrm{e}}_k$ denote the augmented domain-exclusive representations, and $\boldsymbol{z}_k^u=\left(\boldsymbol{Z}_k^{\mathrm{e}}\right)_T$ denote the user representation, where $T$ is the last position of the sequence. Assuming that the batch size is $N$, the augmentation operation yields $2N$ sequences $\{\boldsymbol{z}^{(1)}_k, \widetilde{\boldsymbol{z}}^{(1)}_k, \cdots, \boldsymbol{z}^{(N)}_k, \widetilde{\boldsymbol{z}}^{(N)}_k\}$. Therefore, for each positive pair in the batch, the negative set $\mathcal{Z}^{-}_k$ comprises $2(N-1)$ negative pairs. For example, consider the augmented pair of sequence representations $\boldsymbol{z}^{(1)}_k$ and $\widetilde{\boldsymbol{z}}^{(1)}_k$, the corresponding negative set $\mathcal{Z}^{-}_k$consists of $\{\boldsymbol{z}^{(2)}_k, \widetilde{\boldsymbol{z}}^{(2)}_k,\cdots, \boldsymbol{z}^{(N)}_k, \widetilde{\boldsymbol{z}}^{(N)}_k\}$. Thus, for domain $k$, the intra domain contrastive infomax loss can be formulated as:
\begin{equation}
\setlength{\belowdisplayskip}{8pt}
\begin{aligned}
\mathcal{L}_k^{\text{info}} &=
- I(\boldsymbol{z}^u_k, \widetilde{\boldsymbol{z}}^u_k)\\
&\leq -\mathbb{E}_{\boldsymbol{Z}^\mathrm{e}_k} \log \frac{e^{\left(\text{sim}\left(\boldsymbol{z}_k^{u}, \widetilde{\boldsymbol{z}}_k^u\right) / \tau\right)}}{e^{\left( \text{sim} \left(\boldsymbol{z}_k^u, \widetilde{\boldsymbol{z}}_k^u \right) / \tau\right)}+\sum_{\boldsymbol{z}^{-}_k} e^{\left( \text{sim} \left( \boldsymbol{z}^u_k, \boldsymbol{z}^{\text{-}}_k\right) / \tau\right)}} \\
&\quad\quad\quad+\log \frac{e^{\left(\text{sim}\left(\widetilde{\boldsymbol{z}}_k^u, \boldsymbol{z}_k^{u}\right) / \tau\right)}}{e^{\left(\text{sim} \left(\widetilde{\boldsymbol{z}}_k^u, \boldsymbol{z}_k^u\right) / \tau\right)}+\sum_{\boldsymbol{z}^{-}_k} e^{\left( \text{sim} \left( \widetilde{\boldsymbol{z}}_k^u, \boldsymbol{z}^{\text{-}}_k\right) / \tau\right)}}.
\end{aligned}
\end{equation}
The objective of the formula above is to minimize the distance between the positive sample pair $\boldsymbol{z}^u_k$, $\widetilde{\boldsymbol{z}}^u_k$, and maximize the distance between the negative sample pair $\boldsymbol{z}^u_k$, $\boldsymbol{z}^-_k$ and $\widetilde{\boldsymbol{z}}^u_k$, $\boldsymbol{z}^-_k$, where $\boldsymbol{z}^-_k \in \mathcal{Z}^-_k$.

\subsection{Federated Training and Evaluation}

The overall federated learning algorithm is shown in Algorithm \ref{alg1}. In each round, the server sends the current global model $\boldsymbol{\phi}^{\mathrm{s}, t}$ and global user representations $\boldsymbol{Z}^{\mathrm{g}, t}$ to clients, receives the updated local models $\{\boldsymbol{\phi}^{\mathrm{s}, t+1}\}^K_{k=1}$ and local domain-shared representations $\{\boldsymbol{Z}^{\mathrm{s}, t+1}\}^K_{k=1}$ from clients, and updates the global model using weighted averaging.

In the local training stage, after receiving the global model $\boldsymbol{\phi}^{\mathrm{s}}$ and the global representations $\boldsymbol{Z}^{\mathrm{g}}$, each client updates its local model by the training objectives mentioned above. The overall objective is defined as follows:
\begin{equation}
\begin{aligned}
\mathcal{L}^{\text{total}}_k  & = \mathcal{L}^{\text{disen}}_k + \lambda \mathcal{L}^{\text{info}}_k \\
&= \alpha \mathcal{L}_k^{\text {diff }}+\beta \mathcal{L}_k^{\text {sim }}+\gamma \mathcal{L}_k^{\text {recon }} + \lambda \mathcal{L}^{\text{info}}_k,
\end{aligned}
\label{total loss}
\end{equation}

In the local test stage, we fuse the representations $\boldsymbol{Z}_k^{\mathrm{s}}$ and representations $\boldsymbol{Z}_k^{\mathrm{e}}$, and then we have $\boldsymbol{h}_{k,T}=f_{\theta_k}(\boldsymbol{Z}^{\mathrm{s}}_k + \boldsymbol{Z}^{\mathrm{e}}_k)_T$, where $f_{\theta_k}(\cdot)$ is the prediction layer. Then the probability of being interacted next for item $i$ can be formulated as follows:
\begin{equation}
\begin{aligned}
p_{\boldsymbol{\theta_{k}}}\left(s_{k, T+1}=i \mid \boldsymbol{z}_{k, 1}, \cdots, \boldsymbol{z}_{k, T}\right)=\operatorname{softmax}\left(\boldsymbol{h}_{k, T}\right)_i,
\end{aligned}
\end{equation}
where $\boldsymbol{z}_{k,T}$ is the representation corresponding to the $T$-th time step of the sequence, and $\boldsymbol{h}_{k, T}$ is the output of the $T$-th position of the prediction layer $f_{\boldsymbol{\theta}_k}(\cdot)$.

% \begin{sloppypar}
\begin{algorithm2e}[!ht]
    \caption{FedDCSR}
    \label{alg1}
    \KwIn{Local datasets $\mathcal{D}=\{\mathcal{D}_k\}^K_{k=1}$, local epochs $E$, learning rate $\eta$}
    \KwOut{The optimal encoder parameters $\{\boldsymbol{\phi}_k\}^K_{k=1}=\{(\boldsymbol{\phi}_k^{\mathrm{s}}, \boldsymbol{\phi}_k^{\mathrm{e}})\}^K_{k=1}$}
    % initialization\;
    \textbf{Server executes:} \\
         initialize $\boldsymbol{\phi}^{\mathrm{s}, 0}$\;
         \For{round $t = 0, 1, \cdots T - 1$}{
            \For{each client $k \in K$ in parallel}{
            send $\boldsymbol{\phi}^{\mathrm{s}, t}$ and $\boldsymbol{Z}^{\mathrm{g}, t}$ to client $k$\; 
            % \begin{sloppypar}
            $\boldsymbol{\phi}^{\mathrm{s}, t+1}_{k}, \boldsymbol{Z}^{\mathrm{s}, t+1}_k =~ \textbf{ClientUpdate}(k, \boldsymbol{\phi}^{\mathrm{s}, t}, \boldsymbol{Z}^{\mathrm{g}, t}$);
            % \end{sloppypar}
            }
            $\boldsymbol{\phi}^{\mathrm{s}, t+1} = \sum_{k=1}^{K}\frac{\left|\mathcal{D}_{k}\right|}{\left|\mathcal{D}\right|} \boldsymbol{\phi}^{\mathrm{s}, t+1}_{k}$\;
            $\boldsymbol{Z}^{\mathrm{g}, t+1} = \sum_{k=1}^{K}\frac{\left|\mathcal{D}_{k}\right|}{\left|\mathcal{D}\right|} \boldsymbol{Z}^{\mathrm{s}, t+1}_{k}$;
        }
    \SetKwFunction{FMain}{}
    \SetKwProg{Fn}{ClientUpdate}{:}{}        
    \Fn{\FMain{$k, \boldsymbol{\phi}^{\mathrm{s}}, \boldsymbol{Z}^{\mathrm{g}}$}}{
         save $\boldsymbol{\phi}^{\mathrm{s}}$, $\boldsymbol{Z}^{\mathrm{g}}$ as $\boldsymbol{\phi}^{\mathrm{s}}_k$, $\boldsymbol{Z}^{\mathrm{g}}_k$\;
         % as $\boldsymbol{\phi}^{\mathrm{s}, t}_k$\;
         \For{local epoch $i=1, 2, \cdots, E$}{
            % \tcp{data augmentation}
            % \For{all batch $b \in B$}{
            %     $\Tilde{b} = t(b)$
            % }
            \For{sequence batch $\boldsymbol{S}^{(b)}_k$ from $\mathcal{D}_k$}{
                $\widetilde{\boldsymbol{S}}^{(b)}_k = \text{aug}(\boldsymbol{S}^{(b)}_k)$\;
                $\boldsymbol{Z}^{\mathrm{s}}_k = \text{VGSE}^{\mathrm{s}}_k(\boldsymbol{S}^{(b)}_k; \boldsymbol{\phi}^{\mathrm{s}}_k)$\;
                $\boldsymbol{Z}^{\mathrm{e}}_k, \widetilde{\boldsymbol{Z}}^{\mathrm{e}}_k = \text{VGSE}^{\mathrm{e}}_k(\boldsymbol{S}^{(b)}_k, \widetilde{\boldsymbol{S}}^{(b)}_k; \boldsymbol{\phi}^{\mathrm{e}}_k)$\;
                $
                \begin{aligned}
                \boldsymbol{\phi}^{\mathrm{s}}_k \gets \boldsymbol{\phi}^{\mathrm{s}}_k - \eta \nabla_{\boldsymbol{\phi}^{\mathrm{s}}_k} \Bigl(&\alpha \mathcal{L}^{\text{diff}}(\boldsymbol{Z}^{\mathrm{s}}_k,\boldsymbol{Z}^{\mathrm{e}}_k)\\
                 - &\beta \mathcal{L}^{\mathrm{sim}}\left(\boldsymbol{Z}^{\mathrm{s}}_k,\boldsymbol{Z}^{\mathrm{g}}_k\right)\Bigr)
                \end{aligned}
                $\;

                % \begin{sloppypar}
                $
                \begin{aligned}
                \boldsymbol{\phi}^{\mathrm{e}}_k \gets \boldsymbol{\phi}^{\mathrm{e}}_k
                 - \eta \nabla_{\boldsymbol{\phi}^{\mathrm{e}}_k}\Bigl(&\alpha\mathcal{L}^{\mathrm{diff}}\left(\boldsymbol{Z}^{\mathrm{s}}_k,\boldsymbol{Z}^{\mathrm{e}}_k\right) \\
                 - &\gamma \mathcal{L}^{\mathrm{recon}}\left(\boldsymbol{Z}^{\mathrm{e}}_k,\boldsymbol{S}^{(b)}_k\right)\\
                 - &\lambda \mathcal{L}^{\mathrm{info}}\left(\boldsymbol{Z}^{\mathrm{e}}_k, \widetilde{\boldsymbol{Z}}^{\mathrm{e}}_k\right)\Bigr)
                \end{aligned}
                $;
            }
        }
        \KwRet $\boldsymbol{\phi}^{\mathrm{s}}_k$,$\textbf{ }\boldsymbol{Z}^{\mathrm{s}}_k$;
    }
\end{algorithm2e}
% \end{sloppypar}

\section{Experiments.}
% In this section, we conduct extensive experiments to answer the following research questions:

% In this section, we conduct experiments on sequential recommendation to evaluate the performance of our framework compared with other state-of-the-art methods, we aim to answer the following questions;

% \begin{itemize}
%     \item[\textbf{Q1.}]How does the proposed FedDCSR framework perform compared with the state-of-the-art baselines in the sequential recommendation task(federated version)?
%     % \item[\textbf{RQ2.}]Is FedCSR a general framework that can be applied to various sequential models?
%     % \item[\textbf{RQ3.}]How do different augmentation methods impact the performance? What is the influence of different augmentation hyper-parameters on FedCSR performance?

%     \item[\textbf{Q2.}] How do different components of FedDCSR benefit its performance, i.e., the VGSAN model, disentangled model, and contrastive infomax term?

%     \item[\textbf{Q3.}] Does the domain exclusive and domain shared representations really be able to enhance the representations of cross-domain, single-domain, and unified-domain?

%     \item[\textbf{Q4.}] Does FedDCSR provide a more powerful model and precise cross-domain recommendation results for each data owner?
    
%     % \item[\textbf{Q5.}] Does our FedDCSR really capture the precise cross-domain representations from the user behavior sequence?
    
% \end{itemize}

\subsection{Datasets}

As used in many cross-domain recommendation methods, we utilize the publicly available datasets from Amazon $\footnote{\href{https://jmcauley.ucsd.edu/data/amazon/}{https://jmcauley.ucsd.edu/data/amazon/}}$ (an e-commerce platform) to construct the federated CSR scenarios. We select ten domains to generate three cross-domain scenarios: Food-Kitchen-Cloth-Beauty (FKCB), Movie-Book-Game (MBG), and Sports-Garden-Home (SGH). Following the approach of previous studies \cite{C2DSR, MIFN, PINet}, we remove users and items with fewer than 10 interactions. Additionally, we only preserve sequences that contain a minimum of 4 items, and a maximum of 16 items. For the dataset split, we select the latest 20\% of each user’s interactions as the validation/test set, and the remaining 80\% interaction sequences as the training set. We summarize the statistics of the federated CSR scenarios in Table \ref{table1}.
\begin{table}[!ht]
\vspace{-0.1em}
\renewcommand{\arraystretch}{1.2}
\centering
\Large
\captionsetup{skip=0.1cm}
\caption{\textbf{Statistics of Three Federated CSR scenarios.}}
\label{table1}

\resizebox{\linewidth}{!}{
\begin{tabular}{ccccccccc}

\hline
\textbf{Domain} & \textbf{\#Users} & \textbf{\#Items} & \textbf{\#Train} & \textbf{\#Valid} & \textbf{\#Test} & \textbf{Avg.Len} \\
\hline
Food            & 4658           & 13564  & 4977        & 1307           & 1332 & 8.79   \\
% \hline
Kitchen        & 13382           &     32918                 &             11100                  &        2172 &
2254                          &       8.60          \\ 
% \hline
Clothing          & 9240           & 34909  & 5720            & 818           & 866 & 9.30 \\
% \hline
Beauty    & 5902           &        17780                  &     4668        &      836      &           855            &   10.29
 \\ 
\hline
Movie  & 34792             & 44464 & 57405            & 10944           & 11654 & 7.97 \\
% \hline
Book      & 19419 &   72246                      & 63157            &                          11168 &       12149     &   7.20         &                       & \\
% \hline
Game     & 5588 &    10336               & 6631            &     1374                     &    1444        &   6.49         &                       & \\
\hline
Sport     & 28139 &      88992                   & 51477            &                    13720      &    14214        &   10.65         &                       & \\
% \hline
Garden     & 6852 &    21604                    & 10479            & 3074                         &    3113        &   9.48        &                       & \\
% \hline
Home     & 20784 &   62499                      & 37361            &  10058                        &    10421        &   10.41      &                       & \\
\hline
\end{tabular}}
\end{table}

\begin{table*}[!ht]
\centering
\large
\captionsetup{skip=0.1cm}
% \arrayrulecolor{black}
\caption{\textbf{Federated experimental results(\%) on the FKCB scenario. Avg denotes the average result calculated from all domains. The best results are boldfaced.}}
\label{table2}
\resizebox{\linewidth}{!}{%
\begin{tabular}{cccccccccccccccc} 
\hline
\multirow{3}{0.137\linewidth}{\hspace{10pt}Method} & \multicolumn{3}{>{\hspace{30pt}}m{0.16\linewidth}}{Food}       & \multicolumn{3}{>{\hspace{30pt}}m{0.15\linewidth}}{Kitchen}  & \multicolumn{3}{>{\hspace{30pt}}m{0.16\linewidth}}{Clothing}    & \multicolumn{3}{>{\hspace{30pt}}m{0.154\linewidth}}{Beauty}    & \multicolumn{3}{>{\hspace{30pt}}m{0.154\linewidth}}{Avg}     \\ 
\cline{2-16}
                                                  & \multirow{2}{0.052\linewidth}{\hspace{0pt}MRR} & HR   & NDCG & \multirow{2}{0.05\linewidth}{\hspace{0pt}MRR} & HR   & NDCG & \multirow{2}{0.052\linewidth}{\hspace{0pt}MRR} & HR    & NDCG  & \multirow{2}{0.048\linewidth}{\hspace{0pt}MRR} & HR    & NDCG & \multirow{2}{0.048\linewidth}{\hspace{0pt}MRR} & HR    & NDCG  \\ 
\cline{3-4}\cline{6-7}\cline{9-10}\cline{12-13}\cline{15-16}
                                                  &                                                & @10   & @10  &                                               & @10  & @10  &                                                & @10   & @10   &                                                & @10   & @10  &                                                & @10   & @10   \\ 
\hline
FedSASRec                                 & 6.84                                          & 14.41  & 8.03 & 0.92                                          & 1.64 & 0.95
  & 0.32                                          & 0.46 & 0.33 & 3.66                                            & 7.02  & 4.14 & 2.94                                           & 5.88   & 3.36  \\ 
\hline
FedVSAN                                    & 21.31                                          & 35.21   & 23.92  & 6.46                                           & 12.38 & 7.06 & 1.60                                          & 2.89 & 1.49  & 11.58                                             & 20.70  & 13.01 & 10.24                                            & 17.79   & 11.37  \\
FedContrastVAE                              & 23.38                                           & 39.19 & 26.53 & 7.15                                         & 12.20 & 7.63 & 1.74                                          & \textbf{3.58} & 1.71 & 15.08                                           & 25.73 & 16.88 & 11.84                                           & 20.18  & 13.19  \\ 
\hline

FedCL4SRec                                 & 21.89                                           & 34.53 & 24.32 & 5.56                                          & 9.80 & 5.93 & 1.49                                           & 2.31  & 1.29 & 12.79                                            & 21.52  & 14.27 & 10.43                                           & 17.04  & 11.45  \\ 
% \hline
FedDuoRec                                  & 21.60                                          & 33.63 & 23.93 & 5.45                                          & 9.23 & 5.74 & 1.61                                          & 2.42  & 1.38 & 13.15                                           & 21.05  & 14.45 & 10.45                                           & 16.58  & 11.38  \\

\hline
\textbf{FedDCSR(Ours)}                                     & \textbf{28.87}                                           & \textbf{45.65} & \textbf{32.30}   & \textbf{11.37}                                         & \textbf{21.96} & \textbf{13.07} & \textbf{1.99}                                          & 3.23 & \textbf{1.87} & \textbf{18.73}                                             & \textbf{33.45}  & \textbf{21.66} & \textbf{15.24}                                             & \textbf{26.07} & \textbf{17.23}  \\
\hline
\end{tabular}
}
\end{table*}

\begin{table*}[!ht]
\centering
\small
\captionsetup{skip=0.1cm}
% \large
% \arrayrulecolor{black}
\caption{\textbf{Federated experimental results(\%) on the MBG scenario. Avg denotes the average result calculated from all domains. The best results are boldfaced.}}
\label{table3}
\resizebox{\linewidth}{!}{%
\begin{tabular}{cccccccccccccccc}
\hline
\multirow{3}{0.187\linewidth}{\hspace{30pt}Method} & \multicolumn{3}{>{\hspace{40pt}}m{0.189\linewidth}}{Movie}      & \multicolumn{3}{>{\hspace{40pt}}m{0.175\linewidth}}{Book} & \multicolumn{3}{>{\hspace{40pt}}m{0.189\linewidth}}{Game}   & \multicolumn{3}{>{\hspace{40pt}}m{0.181\linewidth}}{Avg}     \\ 
\cline{2-13}
                                                  & \multirow{2}{0.062\linewidth}{\hspace{0pt}MRR} & HR    & NDCG & \multirow{2}{0.06\linewidth}{\hspace{0pt}MRR} & HR   & NDCG & \multirow{2}{0.062\linewidth}{\hspace{0pt}MRR} & HR    & NDCG  & \multirow{2}{0.056\linewidth}{\hspace{0pt}MRR} & HR    & NDCG  \\ 
\cline{3-4}\cline{6-7}\cline{9-10}\cline{12-13}
                                                  &                                                & @10   & @10  &                                               & @10  & @10  &                                                & @10   & @10   &                                                & @10   & @10   \\ 
\hline
FedSASRec                                   & 10.36                                          & 17.79 & 11.26 & 6.91                                          & 10.91 & 7.19  & 5.20                                          & 6.44 & 5.15 & 7.49                                           & 11.71  & 7.87  \\ 
\hline
FedVSAN                                     & 5.93                                          & 12.72  & 6.43  & 6.54                                          & 13.09 & 7.27 & 1.83                                          & 3.20 & 1.74  & 4.77                                            & 9.67  & 5.15  \\
FedContrastVAE                              & 11.25                                          & 19.14 & 12.22 & 7.63                                         & 12.76 & 8.18 & 4.95                                          & 6.72 & 4.96 & 7.94                                           & 12.87 & 8.45  \\ \hline
FedCL4SRec                                  & 10.39                                           & 17.68 & 11.24 & 6.86                                          & 10.94 & 7.18 & 5.21                                           & 6.44  & 5.17 & 7.49                                           & 11.69  & 7.87  \\ 
% \hline
FedDuoRec                                   & 10.42                                          & 17.83 & 11.55 & 6.79                                          & 10.80 & 7.12 & 5.16                                          & 6.42 & 5.13 & 7.46                                           & 11.68  & 7.93  \\

\hline
\textbf{FedDCSR(Ours)}                                     & \textbf{16.11}                                          & \textbf{28.32} & \textbf{18.09}   & \textbf{10.38}                                         & \textbf{18.68} & \textbf{11.56} & \textbf{7.65}                                          & \textbf{10.60} & \textbf{7.83} & \textbf{11.38}                                             & \textbf{19.20} & \textbf{12.49}  \\
\hline
\end{tabular}
}
\end{table*}

\begin{table*}[!ht]
\centering
\small
\captionsetup{skip=0.1cm}
% \arrayrulecolor{black}
\caption{\textbf{Federated experimental results(\%) on the SGH scenario. Avg denotes the average result calculated from all domains. The best results are boldfaced.}}
\label{table4}
\resizebox{\linewidth}{!}{%
\begin{tabular}{cccccccccccccccc} 
\hline
\multirow{3}{0.187\linewidth}{\hspace{30pt}Method} & \multicolumn{3}{>{\hspace{40pt}}m{0.189\linewidth}}{Sport}      & \multicolumn{3}{>{\hspace{40pt}}m{0.175\linewidth}}{Garden} & \multicolumn{3}{>{\hspace{40pt}}m{0.189\linewidth}}{Home}   & \multicolumn{3}{>{\hspace{40pt}}m{0.181\linewidth}}{Avg}     \\ 
\cline{2-13}
                                                  & \multirow{2}{0.062\linewidth}{\hspace{0pt}MRR} & HR    & NDCG & \multirow{2}{0.06\linewidth}{\hspace{0pt}MRR} & HR   & NDCG & \multirow{2}{0.062\linewidth}{\hspace{0pt}MRR} & HR    & NDCG  & \multirow{2}{0.056\linewidth}{\hspace{0pt}MRR} & HR    & NDCG  \\ 
\cline{3-4}\cline{6-7}\cline{9-10}\cline{12-13}
                                                  &                                                & @10   & @10  &                                               & @10  & @10  &                                                & @10   & @10   &                                                & @10   & @10   \\ 
\hline
FedSASRec                                   & 3.59                                          & 4.34 & 3.43 & 4.74                                          & 5.36 & 4.56  & 3.13                                          & 3.77 & 2.93 & 3.82                                           & 4.49  & 3.64  \\ 
\hline
FedVSAN                                     & 1.16                                          & 2.05  & 1.00  & 1.18                                          & 2.02 & 1.05 & 1.27                                          & 2.26 & 1.10  & 1.21                                            & 2.11  & 1.05  \\
FedContrastVAE                              & 3.93                                          & 4.95 & 3.81 & 5.07                                         & 5.97 & 4.91 & 3.54                                          & 4.37  & 3.39 & 4.18                                           & 5.10 & 4.03  \\\hline 
FedCL4SRec                                  & 3.60                                           & 4.26 & 3.41 & 4.74                                          & 5.27 & 4.55 & 3.11                                           & 3.74  & 2.91 & 3.82                                           & 4.42  & 3.62  \\ 
% \hline
FedDuoRec                                   & 3.75                                          & 4.43 & 3.64 & 4.86                                          & 5.47 & 4.79 & 3.07                                          & 3.60 & 2.83 & 3.89                                           & 4.50  & 3.75  \\
\hline
\textbf{FedDCSR(Ours)}                                     & \textbf{5.29}                                          & \textbf{6.70} & \textbf{5.22}   & \textbf{6.41}                                         & \textbf{8.35} & \textbf{6.52} & \textbf{4.52}                                          & \textbf{5.84} & \textbf{4.45} & \textbf{5.41}                                             & \textbf{6.96}  & \textbf{5.40}  \\
\hline
\end{tabular}
}
\end{table*}
\vspace{-0.5em}
\subsection{Experimental Setting}
\vspace{-0.5em}
\subsubsection{Evaluation Protocol.}

Following previous sequential recommendation works \cite{SASRec, CL4SRec}, we utilize the leave-one-out method to evaluate the recommendation performance. To ensure unbiased evaluation, we adopt the approach described in Rendle's literature\cite{Sample}. Specifically, for each validation or test sample, we calculate its score along with 999 negative items. Subsequently, we evaluate the performance of the Top-K recommendation by analyzing the 1,000 ranked list using metrics like MRR (Mean Reciprocal Rank)\cite{MRR}, NDCG@{10} (Normalized Discounted Cumulative Gain) \cite{NDCG}, and HR@{10} (Hit Ratio).

\subsubsection{Compared Baselines.}

We compare our methods with three types of representative sequential recommendation models: (1) attention-based methods, like SASRec\cite{SASRec}. (2) VAE-based methods, such as VSAN\cite{VSAN} and ContrastVAE\cite{ContrastVAE}. (3) CL-based methods, including CL4SRec\cite{CL4SRec} and DuoRec\cite{DuoRec}. We integrate these methods with FedAvg\cite{FedAvg} to form baselines.

\subsubsection{Implementation and Hyperparameter Setting.}

For all methods, the common hyper-parameters are listed as follows: the training round is fixed as 40, the local epoch per client is fixed as 3, the early stopping patience is fixed as 5, the mini-batch size is fixed as 256, the learning rate is fixed as 0.001, the dropout rate is fixed as 0.3.

\begin{figure*}[ht]
    \centering
    \vspace{-2em}
    \subfigure[Rep. in FKCB scenario]{
    \includegraphics[width=2.1in, scale=0.7]{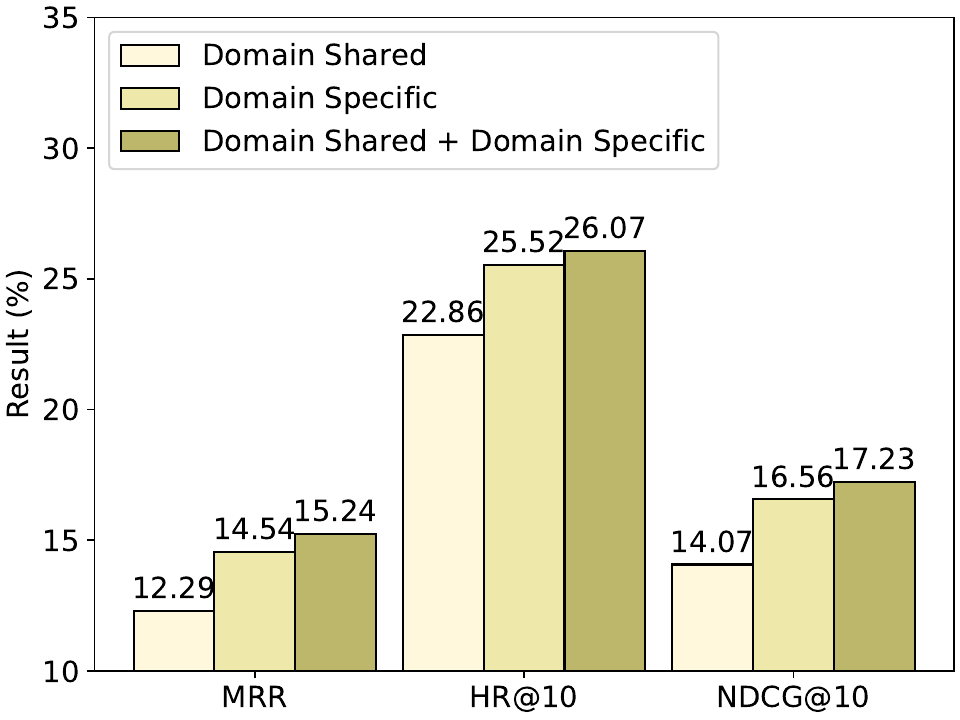}}
    \subfigure[Rep. in MBG scenario]{
        \includegraphics[width=2.1in, scale=0.7]{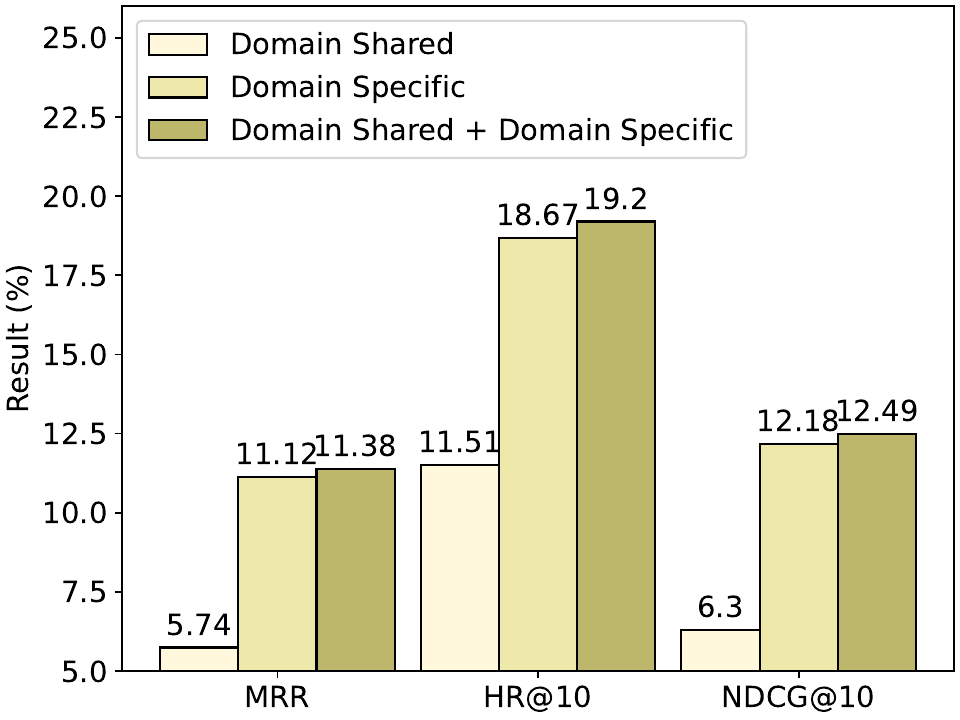}}
    \subfigure[Rep. in SGH scenario]{
        \includegraphics[width=2.1in, scale=0.7]{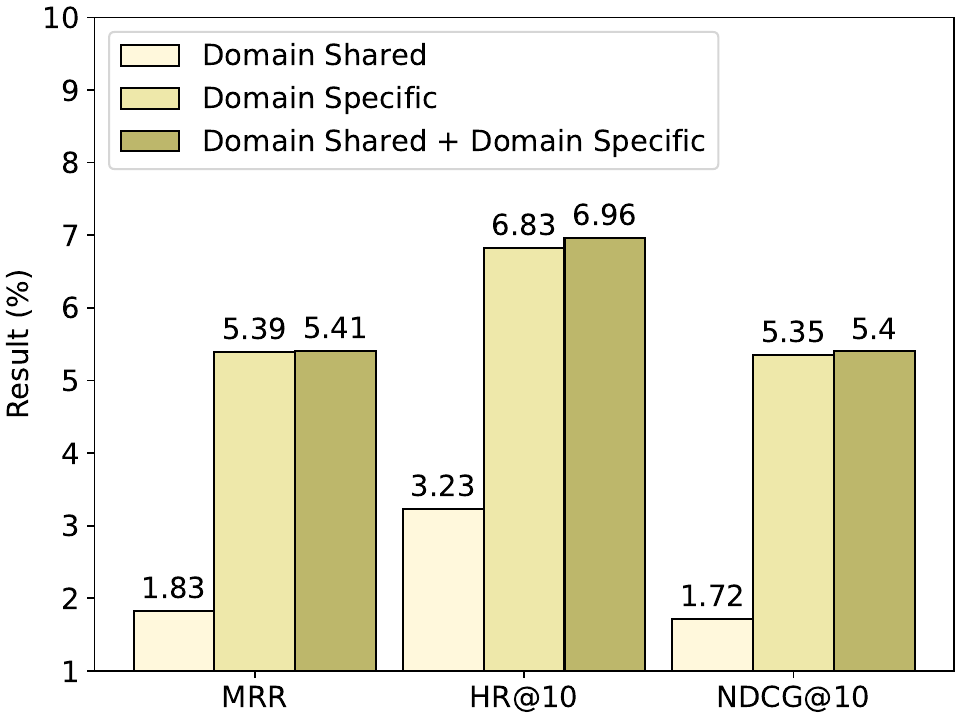}}
    \caption{The predictive results of representations in FKCB, MBG, and SGH scenario.}
    \label{fig5}
\end{figure*}

\subsection{Performance Comparisons}

Table \ref{table2}, \ref{table3}, \ref{table4} show the performance of compared methods on the Food-Kitchen-Clothing-Beuty, Movie-Book-Game and Sports-Garden-Home CSR scenarios.

From the experimental results, we have several insightful observations: (1) For the variational baselines, FedContrastVAE shows better performance than FedSASRec, which validates that modeling the uncertainty of user behaviors could be helpful for better representations in the federated CSR scenario. (2) For the CL baselines, FedCL4SRec and FedDuoRec both show better performance than FedSASRec, which indicates that the contrastive learning has promising advantages in learning representations in the federated CSR scenario. However, we find that CL-based federated CSR sometimes performs equally or worse than FedSASRec, we believe that this is because the CL could intensify the sequence feature heterogeneity and ultimately have a negative impact on the prediction accuracy of the recommendation model. (3) Our FedDCSR significantly outperforms all baselines in many metrics, highlighting the crucial role of disentangled representation learning and contrastive infomax strategy in capturing intra-domain and inter-domain user preferences.

% This fact indicates that utilizing the inter- and intra-sequence item relationships to model the single- and cross-domain user preference is important for CDSR.

% For the cross-domain sequential baselines, $\pi-$Net, PSJNet, MIFN, and $\mathrm{s}^2$DSR reach superior performance to single-domain sequential and cross-domain recommendation baselines, which indicates the cross-domain sequential information is beneficial to enhance recommendation performance. Besides, $\mathrm{s}^2$DSR is the state-of-the-art baseline, which indicates that the contrastive learning and knowledge transfer strategy is also vital to model the cross-domain user representations.
\begin{table}
% \vspace{-1em}
\centering
\large
\caption{Ablation study on FKCB, MBG, and SGH scenarios.}
\label{table5}
\resizebox{\linewidth}{!}{%
\begin{tabular}{cccccccccccccccc} 
\hline
\multirow{3}{0.342\linewidth}{\hspace{0pt}Method} & \multicolumn{2}{>{\hspace{0pt}}m{0.196\linewidth}}{FKCB} & \multicolumn{2}{>{\hspace{0pt}}m{0.19\linewidth}}{MBG} & \multicolumn{2}{>{\hspace{0pt}}m{0.196\linewidth}}{SGH}  \\ 
\cline{2-7}
                                                  & \multirow{2}{0.094\linewidth}{\hspace{0pt}MRR} & NDCG    & \multirow{2}{0.088\linewidth}{\hspace{0pt}MRR} & NDCG      & \multirow{2}{0.094\linewidth}{\hspace{0pt}MRR} & NDCG         \\ 
\cline{3-3}\cline{5-5}\cline{7-7}
                                                  &                                                & @10     &                                                & @10       &                                                & @10          \\ 
\hline
LocalVSAN                                   & 11.57                                          & 13.16    & 5.63                                           & 5.87       & 2.06                                          & 1.96        \\FedVSAN
                                                  &    10.24                                             &  11.37       &   4.77                                             & 5.15          &     1.21                                           &   1.05          \\ 
\hline
LocalContrastVAE                                     & 12.11                                          & 13.38     & 8.67                                           & 9.24      & 4.10                                          & 4.02         \\
FedContrastVAE                                  & 11.84                                           & 13.19    & 8.82                                           & 9.75      & 4.18                                           & 4.03        \\ 
\hline
LocalVGSE                                   & 13.49                                          & 15.30    & 10.63                                           & 11.63      & 5.24                                          & 5.20       \\
FedDCSR - w/o (SRD, CIM)                            & 12.24                                          & 14.01    & 10.31                                          & 11.14      & 5.37                                         & 5.32        \\
FedDCSR - w/o CIM                                    & 14.17                                          & 16.20      & 10.78                                          & 11.67      & 5.40                                          & 5.36        \\ \textbf{FedDCSR(Ours)}
                                                  &        \textbf{15.24}                                        & \textbf{17.23}        &  \textbf{11.38}                                              &      \textbf{12.49}     &    \textbf{5.41}                                            & \textbf{5.40}             \\ 
\hline
\end{tabular}
}
\vspace{-1.5em}
\end{table}

\subsection{Ablation Study}

We perform an ablation study on FedDCSR by showing how the SRD and CIM affect its performance. Table \ref{table5} displays the performance of several model variants in three CSR scenarios. LocalVGSE is the VGSE model without federated aggregation, FedDCSR-w/o (SRD, CIM) is FedDCSR without SRD and CIM, and FedDCSR-w/o CIM is FedDCSR without CIM. We observe that FedDCSR-w/o (SRD, CIM) sometimes performs worse than LocalVGSE, which indicates that the sequence feature heterogeneity is very significant. Additionally, We notice that FedDCSR-w/o CIM performs better than LocalVGSE and FedDCSR-w/o (SRD, CIM), demonstrating the effectiveness of SRD in addressing the feature heterogeneity problem across domains. Finally, CIM allows the model performance to be further improved.

\begin{figure}[!ht]
    \vspace{-1em}
    \centering
    \subfigure[Impact of coefficient $\alpha$]{
    \includegraphics[width=1.6in]{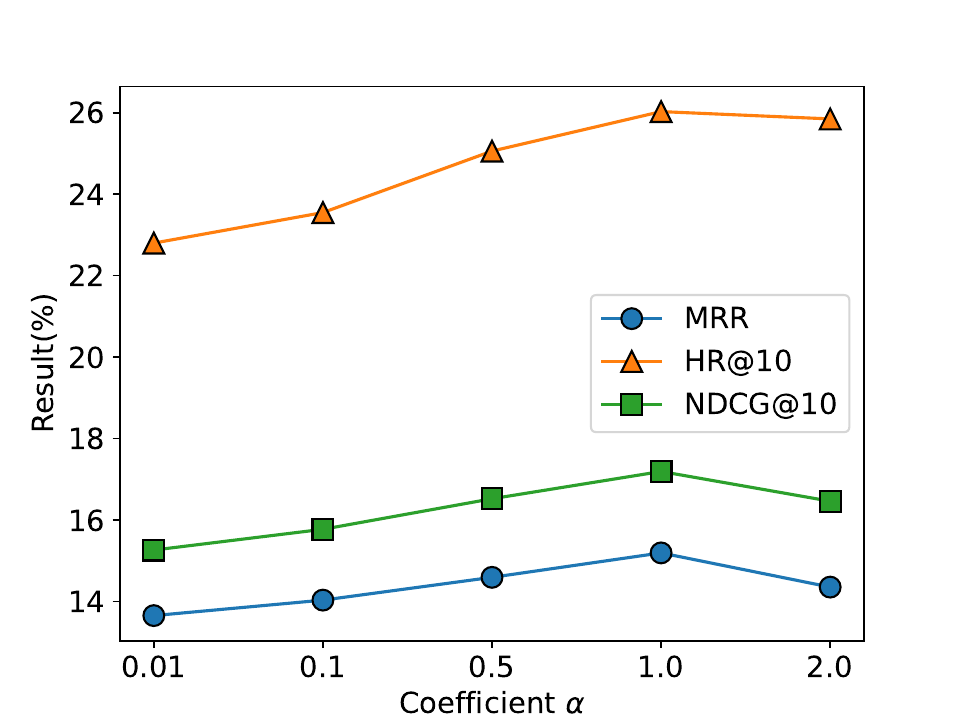}}
    \subfigure[Impact of coefficient $\beta$]{
    \includegraphics[width=1.6in]{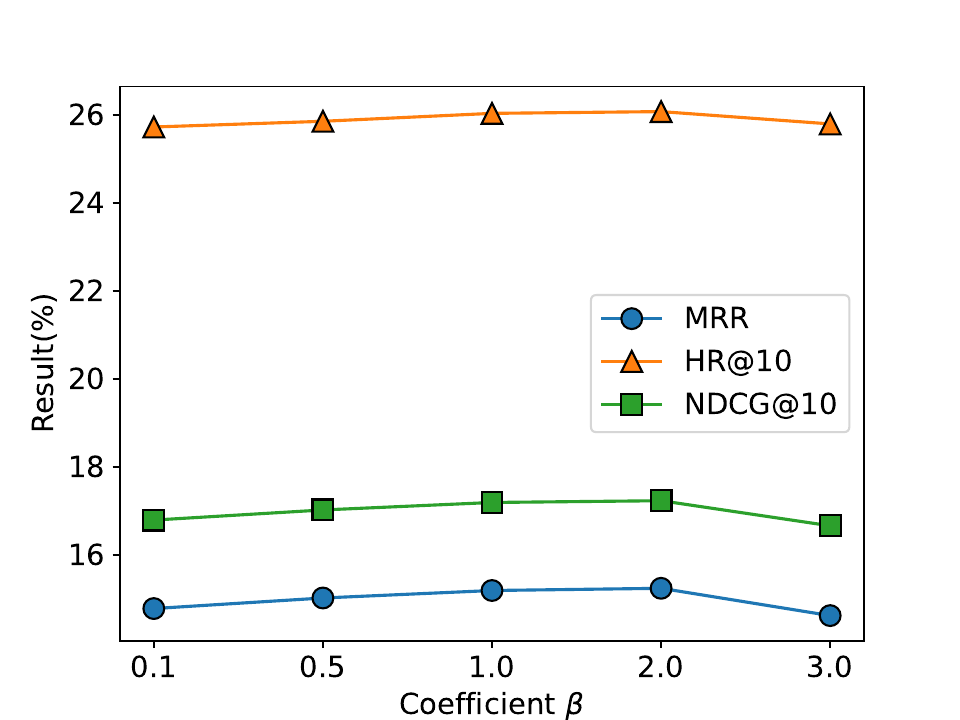}}
    \caption{Impact of coefficient $\alpha$ and $\beta$.}
    \label{fig6}
    \vspace{-2.5em}
\end{figure}

\subsection{Discussion of the user representations}

In this section, to further validate that our SRD can learn domain-shared and domain-exclusive representations for users, we conduct a comparative analysis between the predictive performance of three types of representations: domain-shared, domain-exclusive, and domain-shared+domain-exclusive representations. As shown in Figure \ref{fig5}, our analysis reveals several interesting observations: (1) The predictive results of three types of representations are different, indicating the efficiency of SRD. (2) The domain-shared+domain-exclusive representation outperforms both the domain-shared and domain-exclusive representations, which demonstrates that both leveraging the domain-shared and domain-exclusive features can effectively integrate information from multiple domains.

% resulting in a more comprehensive understanding of user preferences across domains. 
% (3) Our proposed contrastive objective yields improvements in prediction ability for cross-domain, unified-domain, and single-domain representations compared to the model trained without it. We think that this is due to the ability of our contrastive objective to enable the single-domain representations to capture complete user preferences while encouraging the cross-domain representations to provide more generalized preferences across domains.

\subsection{Influence of hyperparameters} Figure \ref{fig6} shows the MRR, @HR10, @NDCG@10 performance with the increase of coefficient $\alpha$ and $\beta$. We can observe that: (1) The overall performance of FedDCSR increases first and then decreases with the increase of $\alpha$, and reaches its peak at $1.0$. It indicates that the $\alpha$ coefficient of $1.0$ is optimal for feature disentanglement. (2) The overall performance of FedDCSR increases first and then decreases with the increase of coefficient $\beta$, and reaches its peak at 2.0. It indicates that a $\beta$ coefficient of 2.0 is optimal for sharing intra-domain information, and shows that there is a trade-off between inter domain and intra domain information.

\section{Conclusion}
In this paper, we propose a novel federated cross-domain sequential recommendation framework FedDCSR, which allows domains to train better performing CSR models collaboratively under privacy protection. Specifically, we propose a sequence representation disentanglement method SRD, which disentangles the user sequence features into domain-shared and domain-exclusive features to address the feature heterogeneity. Besides, we design a contrastive infomax strategy CIM to learn richer domain-exclusive features of users by performing data augmentation on user sequences. 

\section{Acknowledgments}
This work is supported by Key Technologies and Systems of Attack Detection and Situation Assessment Based on Distributed MDATA Model (PCL2022A03-2).

% We have conducted extensive experiments on three real-world cross-domain datasets, and the results show that FedCSR has significantly improved and outperforms the state-of-the-art baselines.

\end{sloppypar}
% \fancyfoot[R]{\scriptsize{Copyright \textcopyright\ 2024 by SIAM\\
% Unauthorized reproduction of this article is prohibited}}
\end{document}